%% file: main.tex
\documentclass[twoside,leqno,twocolumn]{article}

\usepackage[letterpaper]{geometry}

\usepackage{ltexpprt}
\usepackage{hyperref}

\usepackage{algorithm}
\usepackage{algorithmic}

\usepackage{acro}
\usepackage{amsmath}
\usepackage{amssymb}
\usepackage{graphicx}
\usepackage{multirow}
\input{acronyms}

\input{custom_commands}

\begin{document}
\newcommand\relatedversion{}

\title{\Large Spatial-Aware Deep Reinforcement Learning for the Traveling Officer Problem}

\author{Niklas Strauß\thanks{Munich Center for Machine Learning, LMU Munich \{strauss,schubert\}@dbs.ifi.lmu.de}
\and Matthias Schubert\footnotemark[1]}
\date{}

\maketitle

\fancyfoot[R]{\scriptsize{Copyright \textcopyright\ 2024 by SIAM\\
Unauthorized reproduction of this article is prohibited}}

\begin{abstract}\small\baselineskip=9pt The traveling officer problem (TOP) is a challenging stochastic optimization task. In this problem, a parking officer is guided through a city equipped with parking sensors to fine as many parking offenders as possible. A major challenge in TOP is the dynamic nature of parking offenses, which randomly appear and disappear after some time, regardless of whether they have been fined. Thus, solutions need to dynamically adjust to currently fineable parking offenses while also planning ahead to increase the likelihood that the officer arrives during the offense taking place. Though various solutions exist, these methods often struggle to take the implications of actions on the ability to fine future parking violations into account.
This paper proposes SATOP, a novel spatial-aware deep reinforcement learning approach for TOP. Our novel state encoder creates a representation of each action, leveraging the spatial relationships between parking spots, the agent, and the action. Furthermore, we propose a novel message-passing module for learning future inter-action correlations in the given environment. Thus, the agent can estimate the potential to fine further parking violations after executing an action. We evaluate our method using an environment based on real-world data from Melbourne. Our results show that SATOP consistently outperforms state-of-the-art TOP agents and is able to fine up to 22\% more parking offenses.

\end{abstract}

\section{Introduction}
In recent years, \ac{RL} has been successfully applied to tackle complex optimization problems~\cite{bello2016neural,kool2018attention}. One of these optimization tasks that has attracted significant attention is the \ac{TOP}~\cite{shao2017traveling,Shao2019,schmoll2021semi,he2022heterogeneous,zhang2022dynamic} where a parking officer navigates through a sensor-equipped road network and tries to fine as many parking offenders as possible during a shift. Though the locations of current parking offenses are known to the officer due to real-time data from the sensor network, the agent has to travel to the location before writing out a ticket. Thus, the offender might leave before the agent arrives, and new offenses might occur in the meantime.

As pointed out in previous works~\cite{schmoll2021semi,strauss2022reinforcement}, \ac{TOP} yields considerably different challenges from related routing-based problems such as the \ac{VRP} or the \ac{TDP}. For instance, in \ac{TOP}, violations might yield rewards only for a very short time, and offenders do not wait for the agent to arrive. In contrast, \ac{VRP} and \ac{TDP} settings  usually guarantee a certain waiting time, and the cost of not visiting a location is usually considerably higher than in \ac{TOP}. In \ac{TOP}, the agent is not considered to require a significant time to write a ticket and can continue its task from the same location. In contrast, in \ac{VRP} and \ac{TDP}, agents often must spend time handling customers and might have to change location within this time, e.g., in case of a taxi trip.

Several approaches have been proposed to solve \ac{TOP}. These include heuristics~\cite{shao2017traveling}, ant colony optimization~\cite{shao2017traveling}, imitation learning~\cite{Shao2019}, and \ac{RL}~\cite{schmoll2021semi,zhang2022dynamic,he2022heterogeneous}.
However, existing approaches have limitations as they may not fully exploit the spatial relationships between parking locations and the officer's current location, struggle to scale to realistically sized settings, and have difficulty assessing possible future implications of actions due to the dynamic nature of \ac{TOP}. 

In this paper, we propose SATOP, a \ac{RL}-based agent that leverages a novel spatial-aware neural network architecture for \ac{TOP}. Our agent considers all edges containing parking locations as actions and travels on the shortest path to this destination as proposed in~\cite{schmoll2021semi}.
This implies that our agent works on a \ac{SMDP}, as different actions take different times to execute.
To capture the impact of these temporally extended actions, our novel architecture employs a pathing module evaluating the path the agent will take to reach the target location of an action. In addition, we propose a future positioning module, which encodes the correlations between actions and possible future actions using a spatial-aware message-passing mechanism. This way, our new architecture can more easily adapt to the dynamics of \ac{TOP} and learn the potential of fining future offenses. In contrast to existing \ac{RL}-based approaches for \ac{TOP}, our method is able to consistently outperform classical optimization approaches such as ant colony optimization by a large margin. 

In line with previous research, we evaluate SATOP on a simulation environment that replays real-world parking data from Melbourne. We compare our approach with state-of-the-art methods for \ac{TOP}. Our experimental evaluation shows that our method can consistently outperform state-of-the-art approaches, including existing \ac{RL}-based approaches. Furthermore, we investigate the impact of various architectural choices in an ablation study. To summarize, our main contributions are as follows:
\begin{itemize}
\item A pathing module to encode the impact of an action in a more comprehensive way.
\item A novel message passing-based future positioning module encoding inter-action correlations and, thus, the potential of fining future offenses.
\item A joint architecture that outperforms state-of-the-art methods for \ac{TOP} by a significant margin of up to 22\%.
\end{itemize}

The remainder of this paper is organized as follows: Section~\ref{sec:rel_work} provides a comprehensive review of related work, and places \ac{TOP} in the context of similar optimization problems. In Section~\ref{sec:main} and Section~\ref{sec:arch}, we formally define the task and present the architecture of SATOP, including both new modules. Section~\ref{sec:eval} describes our experimental setup and the results of our evaluation. Finally, Section~\ref{sec:conclusion} concludes the paper with a summary of our contributions and ideas for future work.

\section{Related Work}
\label{sec:rel_work}
\subsection{\ac{TOP} and Related Spatial Optimization Problems}
In this section, we compare \ac{TOP} to related problems, namely the \ac{VRP}, the \ac{TSP}, and the \ac{TDP}. We also highlight the unique challenges and differences inherent to \ac{TOP} compared to these related tasks.

The \ac{VRP} has been extensively studied by AI researchers~\cite{gendreau1996stochastic,bello2016neural,Bono2020,kool2018attention,Nazari2018}. It involves one or more vehicles that must visit a given set of customers in minimal time. Though violations and customers represent locations that provide a reward when visited, customers in the \ac{VRP} need to be visited exactly once. In \ac{TOP}, parking spots can dynamically change their status and offenders do not wait for the agent to fine them. Thus, in contrast to \ac{VRP}, agents must reach violations before the offenders leave on their own. 
While various \ac{VRP} variants exist~\cite{Bono2020,kool2018attention,gendreau1996stochastic}, they do not fully include the dynamic and uncertain nature of \ac{TOP}. For instance, some variations consider the appearance of new customers during the day, but they do not include the disappearance of customers after an unknown time interval~\cite{gendreau1996stochastic,Bono2020}. 

The well-known \ac{TSP} can be viewed as a special case of \ac{VRP}. Although one might attempt to simplify \ac{TOP} to a variant of the \ac{TSP} by planning a path through the current violations~\cite{zhang2022dynamic,shao2017traveling}, this approach overlooks a critical aspect: the changing states of parking spots over time. Ignoring this aspect can lead to sub-optimal strategies. Moreover, in \ac{TOP}, the number of parking spots can be significantly higher than in traditional \ac{TSP} instances, making it difficult to solve the \ac{TSP}. In recent years, \ac{RL}-based methods to address the \ac{TSP}~\cite{bello2016neural,kool2018attention} emerged, but the dynamic nature of \ac{TOP} necessitates novel approaches tailored to its unique demands.

The most closely related task to \ac{TOP} is \ac{TDP}, where dispatchers distribute taxis throughout a city to facilitate quick customer pick-ups~\cite{tang2019deep,Kim2021}. However, significant differences exist between \ac{TOP} and \ac{TDP} despite their similarities. In \ac{TDP}, a dispatcher assigns each customer a nearby taxi. In general, the customer is expected to wait for the taxi to arrive, and the dispatcher only assigns taxis that can reach the customer within a maximum waiting time. In \ac{TOP}, the agent has to manage the risk that a violation might disappear before its arrival. Another difference between \ac{TDP} and \ac{TOP} is that after picking up a customer, a taxi has to drive to a location determined by the customer. Thus, information about other available customers is usually outdated at the time of the drop-off. Finally, most \ac{TDP} approaches employ grid abstractions rather than working directly on the road network~\cite{Kim2021}, making them inadequate for handling the fine-grained spatial requirements of \ac{TOP}. Notably, \ac{TDP} typically involves managing a large fleet of taxis, unlike \ac{TOP} tasks that are often executed by a single agent~\cite{tang2019deep,Kim2021}.

\subsection{Methods for Solving \ac{TOP}}
In recent years, various solutions for \ac{TOP} were proposed~\cite{shao2017traveling,schmoll2021semi,zhang2022dynamic,he2022heterogeneous} approaching the problem from different angles.

In~\cite{shao2017traveling}, the authors introduce \ac{TOP} and propose a simple yet effective greedy heuristic that employs a probabilistic model to determine the next parking spot to visit. Additionally, they transform \ac{TOP} into a time-varying \ac{TSP}, which is solved using ant colony optimization. 

Several works have utilized \ac{RL} to tackle \ac{TOP}~\cite{schmoll2021semi,zhang2022dynamic,he2022heterogeneous}. In~\cite{he2022heterogeneous}, the authors introduced a \ac{RL}-based solution for \ac{TOP} using pointer networks inspired by~\cite{bello2016neural}. Furthermore, \ac{TOP} has been approached by using attention-based architectures and \ac{RL} in an attempt to leverage the advantages of attention mechanisms in related tasks like the \ac{TSP}. The authors of~\cite{zhang2022dynamic} adopted the well-known attention-based architecture from~\cite{kool2018attention} to solve \ac{TOP} using \ac{RL}. 
The authors of~\cite{schmoll2021semi} take a distinctive approach by formalizing \ac{TOP} as a \ac{SMDP} with temporally extended actions. They introduce a state encoder combined with \ac{RL} to address \ac{TOP}. Their approach yields comparable performance to previous works in most settings and is only able to outperform them in scenarios with a large number of parking spots. While we follow their formalization with temporally extended actions, we introduce a novel architecture, enabling our approach to consistently outperform all existing baselines in all settings. 

While this paper focuses on the single-agent \ac{TOP}, the \ac{MTOP}, where multiple officers are employed to fine parking violations, has also received attention. A particular challenge in \ac{MTOP} is the coordination among multiple officers. Thus, researchers employed genetic algorithms~\cite{qin2020solving} or \ac{RL} techniques~\cite{strauss2022reinforcement} to tackle the coordination among agents. In~\cite{strauss2022reinforcement}, the authors utilized \ac{RL} to solve the \ac{MTOP}. They adapt the state encoder of~\cite{schmoll2021semi} and introduce several extensions to enable effective officer coordination.

\section{The Traveling Officer Problem}
\label{sec:main}
In \ac{TOP}, an officer tries to fine as many parking offenses as possible by traversing a road network $\graph = (\nodeset, \edgeset, \travelcosts)$, where $\nodeset$ is a set of vertices (intersections), $\edgeset$ is a set of edges (road segments), and $\travelcosts: \edgeset \rightarrow \mathbb{R}^+$ denotes the travel time of the officer for an edge. Each parking spot $\resource \in \resourceset$ is mapped to an edge $\edge \in \edgeset$ of the road network. We refer to the set of parking spots located on edge $\edge$ as $\resourcesonedgeset(\edge)$. The officer observes the state of each parking spot at the current time $t$. The status indicates whether a parking spot is \textit{free}, \textit{occupied}, in \textit{violation}, or already \textit{fined}. 
Whenever the officer passes by a parking spot in violation, the officer writes a ticket, and the spot's status is set to \textit{fined}. 

We model \ac{TOP} as a fixed horizon \ac{SMDP} where the episode length $t_{end}$ corresponds to the duration of the shift of a parking officer. In a \ac{SMDP}, temporally extended actions are introduced, enabling the agent to take actions that span multiple time steps. This allows treating road segments with varying travel times as actions. In addition, we allow the officer to perform extended actions that follow a path from the officer's current location to any edge hosting a parking spot  as in~\cite{schmoll2021semi}.

Formally, a \ac{SMDP} is a five-tuple $(\stateset,\actionset, T,R,\gamma)$, where  $\stateset$ is the set of states, $\actionset$ is the set of actions, $T$ is a probabilistic transition function, $R$ is the reward function, and $\gamma$ denotes the discount factor. Unlike an ordinary \ac{MDP}, the transition function $T$ also comprises the duration of a state transition $\tau$.

We specify the underlying discrete-time \ac{SMDP} for \ac{TOP} as follows:

$\boldsymbol{\stateset}$ represents the set of possible states. Each $\state \in \stateset$ contains information about the current daytime, the officer's location $loc_o$, and the state of each parking spot $\resource \in \resourceset$.

$\boldsymbol{\actionset}$ denotes the set of actions. Following the formulation of~\cite{schmoll2021semi}, the action space corresponds to the subset of all edges $\edgeset$ hosting parking spots. Thus, an action $\action \in \actionset$ corresponds to traveling from the officer's current position to the end of the associated edge $\edge_\action \in \edgeset$. The agent's path can be computed with a dedicated policy. Here, we compute the shortest path w.r.t.\ the agent's travel time and compute paths with Dijkstra's algorithm. We denote the set of parking spots along the route from the officer's current position to the target of action $\action$ as $\resourcesonrouteset_\action$~\footnote{For the sake of readability, we omit the current location officer $loc_o$ from our notation.}. 

$\boldsymbol{T}(\state_{t+1}, \tau | \state_t, \action_t):  (\stateset \times \mathbb{R}^+ \times \actionset \times \stateset) \rightarrow [0, 1]$ denotes the state transition probabilities. While the agent's position change is deterministic, the state  of parking spots is non-deterministic. A parking spot's state might change due to the officer fining an offense or an unknown stochastic process modeling parking occupancy. The transition function also includes the duration of the transition $\tau$, given by the travel time along the path representing action $\action$.

$\boldsymbol{R}: (\stateset \times \actionset \times \stateset) \rightarrow \mathbb{R}$ denotes the reward function. The agent receives a time-discounted reward $\zeta_t = \sum_{j=0}^{\tau-1}{\gamma^j} r_{t+j+1}$, where $r_{t+j}$ denotes the reward received at time step $t+j$ during the temporally extended action $\action$ with duration $\tau$ time-steps. The officer gets a reward of +1 for any parking violation fined along the path corresponding to action $\action$.

The objective of an agent is to maximize the expected time-discounted reward $\mathbb{E} \left[ \sum_{t=0}^{t_{end}} \gamma^t r_{t+1} \right]$.

\section{A Spatial-Aware TOP Agent}
\label{sec:arch}

\begin{figure}[t]
    \centering
    \includegraphics[width=1.0\columnwidth]{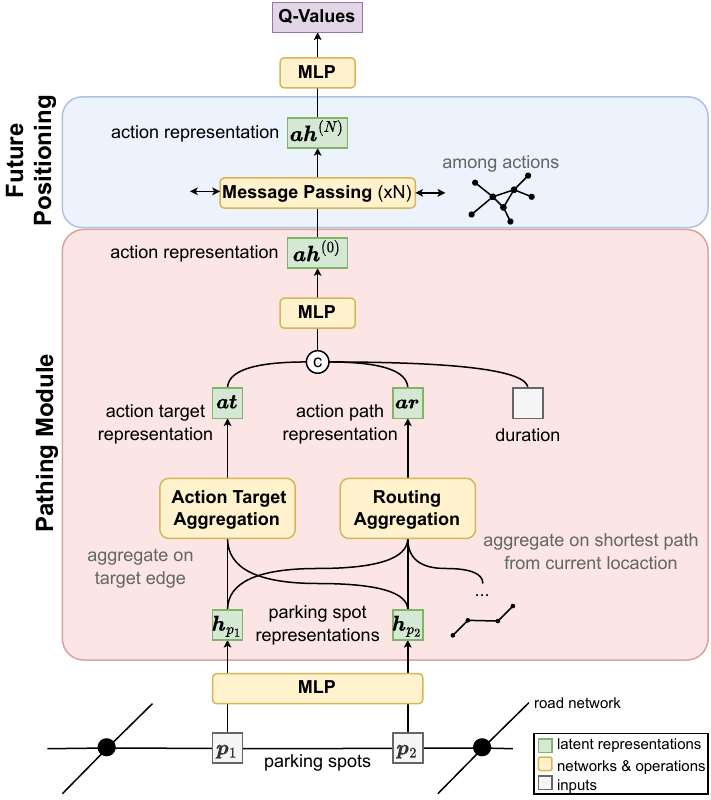}
    \caption{A schematic overview of our novel architecture consisting of the pathing module (red) and the future positioning module (blue). Best viewed in color.}
    \label{fig:arch}
\end{figure}

A good policy for \ac{TOP} requires the agent to consider two aspects: During the current action, the agent has to anticipate the likelihood that parking spots are in violation upon arrival. In addition, the agent has to assess the potential to fine further parking violations during future actions. Our novel architecture effectively covers both aspects by incorporating the spatial relationships between parking locations and the agent's location with a novel pathing module and a future positioning module to assess the potential to fine parking violations within future actions. A schematic overview of our novel architecture, named SATOP, can be found in Figure~\ref{fig:arch}. For readability, we detail the exact parameters of our architecture in the Appendix.

\subsection{Input Features}
Even though the agent observes the current status of the parking spots and its own position, this observation does not satisfy the Markov property. For instance, information about the current parking duration and the maximum allowed parking duration yield essential information for learning transition probabilities. Therefore, as in ~\cite{schmoll2021semi}, we enrich the observation with additional features summarizing the relevant history. Let us note that we explored training LSTM and Transformer-based encoders to learn those features. However, using the handcrafted features generated comparable results while being significantly more computationally efficient.
The input to our agent consists of a feature vector for all parking spots $\resource_i \in \resourceset$ describing the state of $\resource_i$ relative to the agent's position $loc_o$ and the current time of the day. The input vector for a parking spot $\resource_i$ at time $t$ contains its current status, either \textit{free}, \textit{occupied}, \textit{violation}, or \textit{fined}, as a one-hot encoding. 
In addition, we add $x$ and $y$ coordinates of the spot's location being normalized to the operation area's bounding rectangle, which helps identify spatially related parking opportunities. Furthermore, we add a normalized timestamp to allow the agent to differentiate between daytimes. To estimate the likelihood of a status change, we provide an additional feature indicating the remaining parking duration for occupied spots. The same feature encodes the time interval a parking spot yields a violation if its status is \textit{violation}. In particular, a value of -1 to 0 indicates the remaining allowed parking time, while a positive indicator up to a value of 2 indicates how long the parking spot is already in violation.  
In addition, we provide a feature indicating whether an occupied parking spot might exceed the maximum allowed parking duration within the time the officer would need to reach this parking spot. This way, the agent can learn to travel toward parking spots that are not in violation yet but might be soon. Finally, we add features for the officer's normalized distance, travel time, and arrival time to the parking spot $\resource_i$.

\subsection{Parking Spot Representations}
First, we employ an MLP to generate a latent $d_{\resourcerep}$-dimensional representation $\boldsymbol{\resourcerep}_{\resource_i}$ for each parking spot $\resource_i \in \resourceset$ based on its current observation and a $d_{le}$-dimensional learnable embedding vector. While the weights of the MLP are shared across all parking spots, the embedding vector is not shared to allow the network to learn parking spot-specific representations. 

\subsection{Pathing Module}
In our formulation of \ac{TOP}, each action $a$ corresponds to traveling to a specific action target $\edge_a$. Since the agent's current position varies, the parking spots the agent will pass by when executing an action also vary. Therefore, we compute an encoding for each action $a$, which captures the information of parking spots along the corresponding path to the action target $e_a$. All modules in our architecture described in the following either aggregate information to a per-action level or work on per-action representations. The weights of the modules are shared across all actions. To enhance readability, we describe our architecture on a per-action level. In practice, implementations perform the computations for all actions simultaneously using matrix operations. Our novel pathing module consists of several components described in the following.

\textbf{Action Target Encoder}
To describe an action $a$ in our setting, we first create a representation of each action target $e_a$ by building a weighted aggregation over the parking spots $\resource_i \in PE(\edge_a)$. The action target $e_a$ yields essential information since it represents the agent's position after action $a$ is performed. Thus, it plays an important role for the future positioning module described in Section \ref{subsec:future_pos_module}.
In particular, we combine the latent representation $\boldsymbol{\resourcerep}_{\resource_i}$ of the parking spots $\resource_i \in \resourcesonedgeset(\edge_a)$ on each action's $a$ target edge $\edge_a$ into a vector $\boldsymbol{at}_a$ in the following way:
\begin{equation}
    \boldsymbol{at}_a = \sigma \bigl( \bigl( \sum_{\resource_i \in \resourcesonedgeset(\edge_a)} W^{at} \boldsymbol{\resourcerep}_{\resource_i} \bigr) + \boldsymbol{b}^{at} \bigr),
\end{equation}
where $W^{at}$ ($d_{\resourcerep} \times d_{at}$) and $\boldsymbol{b}^{at}$ denote the parameters of the layer, which are shared among all actions, and $\sigma(\cdot)$ is the activation function.

\textbf{Route Aggregation Module}
In addition, we introduce the route aggregation module, where we aggregate the latent representations of parking spots along the path from the officer's current location $loc_o$ to the action target $\edge_a$. In this aggregation, we weight the latent parking spot representations $\boldsymbol{\resourcerep}_{\resource_i}$ by the normalized travel time $\hat{\phi}_{a}(loc_o, \resource_i)$ from the officer's current location $loc_o$ to the parking spot $\resource_i \in \resourcesonrouteset_a$ along the route to the action target  $\edge_a$. Additionally, we scale the normalized travel times using a learnable parameter $\theta \in \mathbb{R}$ that is shared among all actions. This results in a $d_{\resourcerep}$-dimensional path representation $\boldsymbol{ar}_a$ for each action $a$ that is computed as follows:
\begin{equation}
    \boldsymbol{ar}_a = \sum_{\resource_i \in \resourcesonrouteset_a} \theta \cdot \hat{\phi}_{a}(loc_o,\resource_i) \cdot \boldsymbol{\resourcerep}_{\resource_i}
\end{equation}

\textbf{Action Representations}
Next, we concatenate the path representation $\boldsymbol{ar}_a$ with the corresponding action target representation $\boldsymbol{at}_a$  and the expected travel time to the action target $duration_a$\footnote{Let us note that $duration_a$ is the computed travel time for the shortest path the agents computed, whereas $\tau$ describes the required time of an action provided by the environment.} for each action $a$. 

We pass this combined information through an MLP, resulting in $d_{ah}$-dimensional action representations $\boldsymbol{ah}_a^{(0)}$ which combines all three inputs:
\begin{equation}
    \boldsymbol{ah}_a^{(0)} = \text{MLP}_{ah}([\boldsymbol{ar}_a, \boldsymbol{at}_a, duration_a])
\end{equation}
Here, $[\cdot,\cdot,\cdot]$ is the action-wise concatenation operator, and the MLP parameters are shared between all actions. 

\subsection{Future Positioning Module}
\label{subsec:future_pos_module}
To estimate how well the agent is positioned after each action, we perform message passing between actions and possible future actions. In this process, the agent is able to aggregate information from potential future actions into each action's representation utilizing the spatial relationship between them. Therefore, we create a graph structure where nodes contain the latent representations of actions and edges link to possible future actions.

We already created rich representations $\boldsymbol{ah}_a^{(0)}$ for each action ("node") in the pathing module.
To compute information about the links between actions, we define the edge information between action $a$ and a possible future action $a'$ as $\boldsymbol{\delta}_{a,a'}$. In particular, we employ the travel time and the number of parking spots along the path between $\edge_a$ and $\edge_{a'}$ as $\boldsymbol{\delta}_{a,a'}$. 
Then, we transform each edge information $\boldsymbol{\delta}_{a,a'}$ into an importance factor $\hat{\delta}_{a,a'}^{(l)}$ by passing the information through an MLP with an output size of 1, followed by a tanh activation to constrain its range:
\begin{equation}
    \hat{\delta}_{a,a'}^{(l)} = \text{tanh} \bigl( \text{MLP}_{\delta}^{(l)}(\boldsymbol{\delta}_{a,a'}) \bigr)
\end{equation}
The parameters of this MLP are shared across all links.

In the next step, we use this importance factor to combine the information $\boldsymbol{ah}_{a'}^{(l-1)}$ from all possible future actions $a' \in A$ for each action $a$:
\begin{eqnarray}    
 & &  \boldsymbol{\hat{ah}}_a^{(l)} = \sigma \bigl( \bigl( \sum_{{a'} \in A} \hat{\delta}_{a,a'}^{(l)} W^{ah,(l)} \boldsymbol{ah}_{a'}^{(l-1)} \bigr) + \boldsymbol{b}^{ah,(l))} \bigr) \\
 & & \boldsymbol{ah}_a^{(l)} = \text{LN}^{(l)} \bigl( \boldsymbol{\hat{ah}}_a^{(l)}  + \boldsymbol{ah}_{a}^{(0)}   \bigr)
\end{eqnarray}
Here, $l \in \{1, \dots, N \}$ denotes the layer index. We share the $(d_{ah} \times d_{ah})$ weight matrix $W^{ah,(l)}$ and $\boldsymbol{b}^{ah,(l)}$ between all actions and possible future actions but not between layers. For each layer, we apply layer normalization (LN)~\cite{ba2016layer} and incorporate residual connections~\cite{he2016identity}. We choose ELU~\cite{clevert2015fast} as the activation function $\sigma(\cdot)$.

We use two future positioning module layers to enable efficient message passing and parking spot information aggregation across possible future actions. The process ensures that the neural network captures significant spatial and temporal dependencies between possible future actions and parking spots.

\subsection{Q-Value Estimation}
Finally, we employ another MLP with an output size of 1 to reduce the action representation $\boldsymbol{ah}_a^{(N)}$ of each action $a$ into a final Q-Value, which provides estimates of the expected return associated with each action:
\begin{equation}
    Q_a = \text{MLP}_{Q}(\boldsymbol{ah}_a^{(N)})
\end{equation}
The weights of the network are shared between actions.

Employing our proposed neural network architecture enables the agent to effectively assess parking violations on the route and anticipate its future positioning after executing actions in \ac{TOP}, thus allowing the agent to learn effective policies.

\subsection{Training}
Given the formulation of \ac{TOP} with temporally extended actions, we utilize DoubleDQN~\cite{van2016deep} adapted to the semi-Markov setting to train the agent. While several state-of-the-art \ac{RL} algorithms are based on policy gradients~\cite{sutton1999policy,schulman2017proximal,haarnoja2018soft}, their application to our architecture is difficult. Many of these algorithms require learning a compatible shared representation that actor and critic utilize~\cite{sutton1999policy}. However, this proved challenging with our architecture because we separated the representation of different actions early on. To still explore these approaches, we conducted experiments training our proposed architecture using algorithms like PPO~\cite{schulman2017proximal} and SAC~\cite{haarnoja2018soft} on a concatenation of all action representations. Unfortunately, these attempts resulted in sub-optimal performance and training instabilities.
Hence, we turned our attention to the DoubleDQN algorithm, which has successfully dealt with \ac{TOP}~\cite{schmoll2021semi}.
We utilize a replay buffer and update the parameters of our network using a batch of transitions. We minimize the following loss function:
\begin{equation}
    \mathcal{L}(\Theta) = \mathbb{E}_{\state_t, \action_t, \tau, r_{t:t +\tau -1}} \left[(y_t - Q(\state_t ,\action_t; \Theta) )^2 \right]
\end{equation} where $y_t = \sum_{j=0}^{\tau -1} \gamma^j r_{t+j} + \gamma^{\tau} Q(\state_{t+\tau}, \action'_{t+\tau}; \Theta')$. Here, $\action'_{t+\tau}$ represents the optimal action with respect to $\Theta$, i.e., $\action'_{t+\tau} = \argmax_{\action_{t+\tau} \in \actionset(\state_{t+ \tau})} Q(\state_{t+\tau}, \action_{t+\tau}; \Theta)$. $\Theta$ corresponds to the parameters of the behavior $Q$ network and $\Theta'$ represents the parameters of the frozen target $Q$ network, which are periodically updated by copying from $\Theta$.

\section{Evaluation}
\label{sec:eval}

\begin{figure}[t]
    \centering
    \includegraphics[width=0.6\columnwidth]{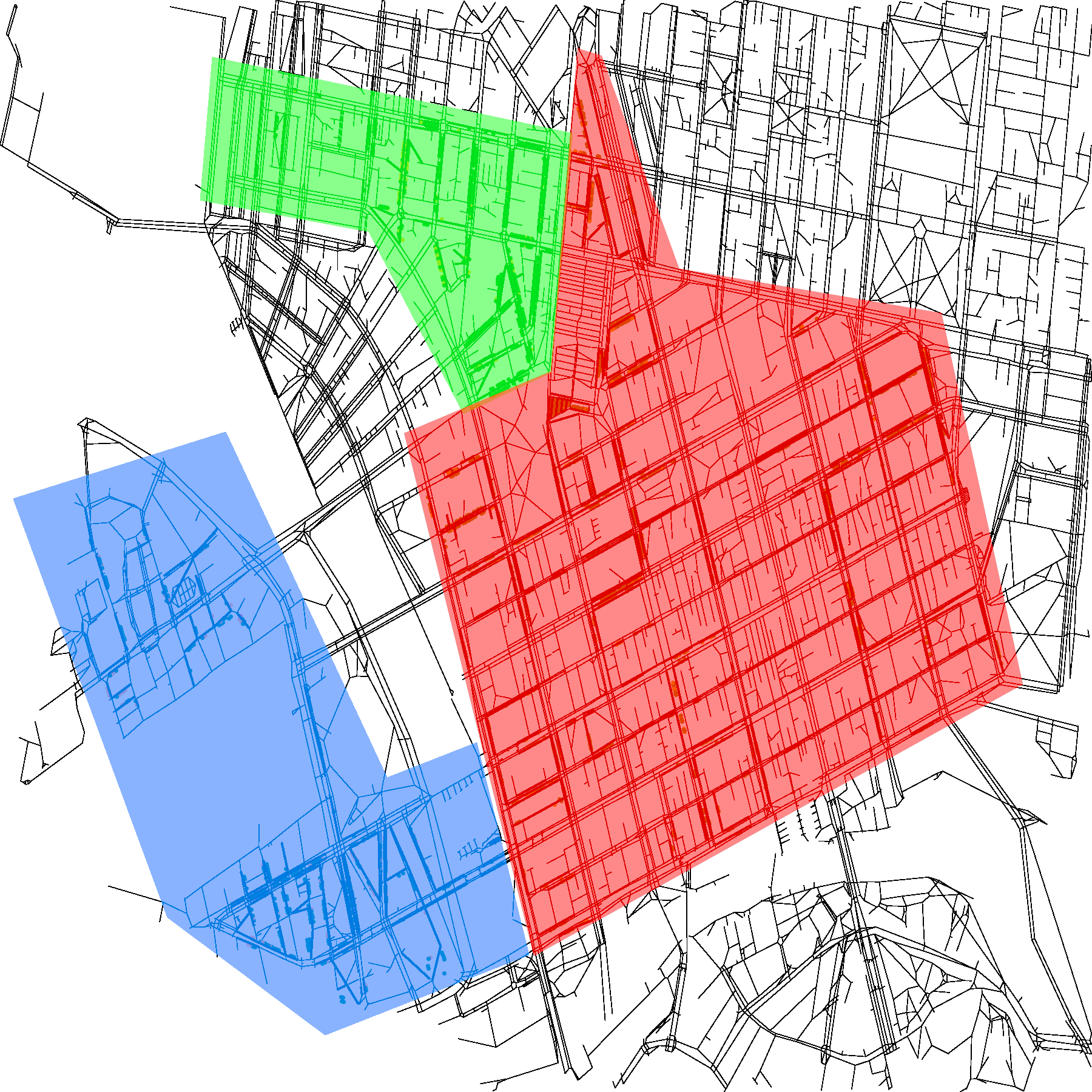}
    \caption{Visualization of the areas used in our evaluation: Queensberry (green), Docklands (blue), and Downtown (red). Best viewed in color.}
    \label{fig:map}
\end{figure}

In this section, we provide an extensive experimental evaluation of our novel approach, SATOP.
We implement an event-based simulator in C++ for \ac{TOP} with OpenAI Gym-compatible Python bindings. To replicate real-world conditions, we utilize parking data from Melbourne in 2019, sourced from the city's open data platform~\footnote{\url{https://data.melbourne.vic.gov.au/explore/dataset/on-street-car-parking-sensor-data-2019/information/}}. To create different graph structures and explore the transferability of hyperparameters, the city of Melbourne is divided into three distinct areas, namely Docklands, Queensberry, and Downtown. We obtain the walking graph from OpenStreetMap~\footnote{\url{https://www.openstreetmap.org}}. The characteristics of these areas are summarized in Table~\ref{tab:areas} and Figure~\ref{fig:map}.
To ensure an unbiased evaluation, we split the parking event dataset into a training, validation, and test set. Since parking patterns tend to exhibit weekly trends and to avoid biases introduced through weekdays, we account for this by partitioning the dataset by the remainder of dividing the day of the year by 13: If the remainder is 0, the day is included in the test set. If the remainder is 1, we add the day to the validation set. The remaining days are assigned to the training set. We consider each day as an episode and to improve diversity, we shuffle the order of the training days. The officer's workday spans 12 hours each day from 7AM to 7PM. We set the travel speed of the officer to 5km/h. To accelerate the training process, the agent interacts with several environments simultaneously. We implement our approach within the Tianshou framework~\cite{tianshou}, which provides a reliable and efficient platform for \ac{RL}. Hyperparameters were tuned using the Docklands area and the resulting hyperparameters were applied to all areas. The test results are generated by using the weights achieving the highest validation results during training. Each training run is executed on a single GPU within a cluster consisting of various GPUs equipped with 24GB to 48GB of GPU memory. The full details regarding the environment parameters, hyperparameters, and training procedures can be found in the Appendix~\footnote{\label{githubfootnote}\url{https://github.com/niklasdbs/satop}}\addtocounter{footnote}{-1}\addtocounter{Hfootnote}{-1}.
As a significant part of the research in this domain does not publish their implementation, uses data from different years, and employs varying pre-processing and data splits, our work strives to provide an extensive comparison to related approaches and establish a benchmark in the field. To promote standardization and facilitate future research, we openly publish our framework and baseline implementations~\footnotemark.

\begin{table}[t]
    \centering
    \begin{tabular}{|c||c|c|c|}
        \hline
        Area            & Docklands & Queensberry & Downtown \\ \hline \hline
        Nodes      & 1,435 & 1,711 & 6,806 \\ \hline
        Edges      & 4,307 & 5,356 & 21,369\\ \hline
        P. Spots & 487   & 639   & 1,481 \\ \hline
        Actions    & 166   & 177   & 493   \\ \hline
    \end{tabular}
    \caption{Characteristics of the different areas used in our evaluation. Note that the number of actions is much smaller than the number of parking spots or edges.}
    \label{tab:areas}
\end{table}

\subsection{Baselines}
\begin{table*}[t]
    \centering
\begin{tabular}{|l||c|c||c|c||c|c| }
\hline
 \multicolumn{1}{|r||}{Area}                    & \multicolumn{2}{c||}{Queensberry} & \multicolumn{2}{c||}{Docklands} & \multicolumn{2}{c|}{Downtown} \\
 Algorithm              &  Validation & Test & Validation & Test & Validation & Test  \\   \hline \hline
 Greedy~\cite{shao2017traveling}                 &  166.59 & 166.79 &     223.59 & 224.96 & 209.48 & 208.96 \\ \hline
 ACO (0.1s)~\cite{shao2017traveling}             &  192.17 & 192.36 &     262.45 & 261.89 & 231.86 & 231.89 \\ \hline
 ACO (1.0s)~\cite{shao2017traveling}             &  197.21 & 196.93 &     271.24 & 271.11 & 261.66 & 253.57 \\ \hline
 SDDQN~\cite{schmoll2021semi}                  &  185.31 & 187.07 &     268.59 & 267.25 & 292.41 & 289.46 \\ \hline
 DGAT~\cite{zhang2022dynamic}                   &  159.1  & 158.64 &     181.28 & 176.46 & 138.59 & 137.18 \\ \hline
 PTR~\cite{he2022heterogeneous}                    &  161.31 & 159.04 &     223.55 & 222.11 & 150.41 & 149.39 \\ \hline
 \textbf{SATOP (ours)}  &  \textbf{209.45} & \textbf{210.89} & \textbf{314.31}  & \textbf{310.71} & \textbf{359.0} & \textbf{353.25} \\ \hline

\end{tabular}

\caption{Average number of parking violations fined per day across Queensberry, Docklands, and Downtown on the validation and test data.}
\label{table:results}
\end{table*}

In order to evaluate the performance of our proposed neural network architecture SATOP, we compare it against several state-of-the-art methods commonly used in this domain.

\subsubsection{Greedy}
The greedy approach, proposed by~\cite{shao2017traveling}, uses a probability model to determine the next parking spot to visit. It selects the spot with the highest probability of still being in violation upon arrival. The approach assumes that the distribution of parking violations follows an exponential distribution. While this method provides a simple and easy-to-compute solution, it yields surprisingly effective results. 

\subsubsection{ACO}
The authors of~\cite{shao2017traveling} address \ac{TOP} by simplifying it to time-varying \ac{TSP}. By applying ant colony optimization, this approach attempts to find an optimal sequence of visits to different locations. 
As the ACO approach requires computing the solution during inference, we limit the computation time to 0.1 and 1.0 seconds.

\subsubsection{SDDQN}
\cite{schmoll2021semi} developed a state-of-the-art \ac{RL} approach for \ac{TOP}. Its performance serves as a benchmark for evaluating the effectiveness of our novel architecture.

\subsubsection{PTR}
The authors of~\cite{he2022heterogeneous} propose simplifying \ac{TOP} to a variant of the \ac{TSP} and solve it using pointer networks based on the work of~\cite{bello2016neural}.

\subsubsection{DGAT}
In~\cite{zhang2022dynamic}, the authors apply the well-known approach from~\cite{kool2018attention} to solve \ac{TOP} using an attention mechanism. They also simplify \ac{TOP} to a variant of the \ac{TSP}.

\subsection{Metrics}
The primary objective of \ac{TOP} is to fine as many parking violations as possible. Hence, we use the average number of fined violations per day as our primary metric for assessing performance. 

\subsection{Results}

\begin{table*}[t]
    \centering
    \begin{tabular}{|l || c|c || c|c |}
    \hline
    \multicolumn{1}{|c||}{Area}                                & \multicolumn{2}{c||}{Docklands} & \multicolumn{2}{c|}{Downtown} \\
    Ablation                            & Validation        & Test              & Validation        & Test  \\   \hline \hline
    SATOP (ours)                        & \textbf{314.31}   & \textbf{310.71}   & \textbf{359.0}    &  \textbf{353.25} \\ \hline
    No Future Pos. Module               & -31.24            & -30.07            & -32.72            & -34.82 \\ \hline
    Future Pos. Module: Only Travel Time as Link Info ($\delta$) & -3.76       & -1.14             & -4.76             & -7.79 \\ \hline
    Pathing Module: No Action Target Encoder            & -6.10             & -5.57             & -6.62             & -11.36 \\ \hline
    Pathing Module: No Route Agg. Module                & -3.38             & -5.21             & -7.07             & -9.07 \\ \hline
    No Spot Specific Representations    & -5.93             & -5.53             & -4.38             & -7.64 \\ \hline
    
    \end{tabular}
    \caption{Ablations evaluated on the Docklands and Downtown area. We report the reduction in average number of fined violations per day.}
    \label{tab:ablation}

\end{table*}

In this section, we demonstrate the superior results of SATOP across different areas, namely Queensberry, Docklands, and Downtown. The performance of various approaches including Greedy, ACO with different time limits (0.1s and 1.0s), SDDQN, PTR, DGAT, and our proposed approach SATOP are evaluated based on both validation and test sets. 

Table~\ref{table:results} displays the average number of fined parking violations per day for each approach in the different areas. The results demonstrate the outstanding performance of our approach, outperforming all other algorithms across all areas on both the validation and test data. 

In the Queensberry area, SATOP achieves a significantly better average number of fined violations with a score of 210.89 (test set) and 209.45 (validation set). It outperforms all other methods, with DGAT, PTR, and Greedy trailing far behind with test scores of 158.64, 159.04, and 166.79, respectively. SDDQN achieves a score of 187.07. While ACO with different time limits shows promising results with test scores of 192.36 (0.1s) and 196.93 (1.0s), SATOP is still significantly more effective in this scenario. 

In the Docklands area, SATOP continues to show its superiority, achieving a score of 314.31 on the validation set and 310.71 on the test set, easily surpassing all other methods. The closest competitor in this area is the ACO (1.0s), with a score of 271.24. In line with previous research, SDDQN (with a score of 267.25) is unable to outperform the ACO, but it still manages to outperform the Greedy baseline (224.96) significantly. 

In the Downtown area, our approach again stands out as the top-performing method, with a score of 353.25 (test set) and 359.0 (validation set). Downtown is the largest and most challenging area. Notably, DGAT and PTR trail far behind in this scenario and exhibit a drop in performance compared to Docklands. A possible explanation for this is that they treat \ac{TOP} as a variant of the \ac{TSP}, which is difficult to solve because of the large number of parking spots. ACO (1.0s) also shows a slight decrease in performance (261.66) compared to Docklands (271.24), while SATOP manages to fine significantly more parking violations than in Docklands. SDDQN, our closest competitor, is able to outperform all other baselines with scores of 289.46 (test set) and 292.41 (validation set). Still, our approach is able to fine nearly 64 more parking violations per day on average on the test data, which is an improvement of 22\%.

Furthermore, our results reveal that DGAT and PTR perform inferior in all areas. These approaches treat \ac{TOP} as a variant of the \ac{TSP} and solve it using \ac{RL}. In both papers, the authors limited the evaluation to only a small number of parking spots (150 max), which is only 10\% of the parking spots in Downtown and around 30\% of the parking spots in Docklands (the area with the fewest parking spots).  Our results indicate that these approaches have difficulties scaling to real-world scenarios. Furthermore, this underscores the distinct characteristics of \ac{TOP} and the \ac{TSP}.

Overall, the results presented in Table~\ref{table:results} provide compelling evidence of the effectiveness of SATOP. Our method consistently outperforms other algorithms across all areas, demonstrating its potential to effectively solve \ac{TOP} in various settings.

\subsection{Ablations}
We conducted various ablations to assess the influence of the different components within our novel architecture. These ablations were carried out in the Docklands and Downtown area and tested on both the validation and test datasets. To measure the performance, we use the average number of violations fined per day. The results of our ablation study are presented in Table~\ref{tab:ablation}. Notably, the outcomes of the ablations are consistent across different areas and both the validation and test datasets.

Removing the future positioning module has a substantial impact on performance. In both Docklands and Downtown, eliminating this module resulted in a sharp reduction of approximately 30 fewer parking violations fined per day. This performance drop was consistent across both the test and validation datasets. In our future positioning module, we make use of complex edge data. Using only the travel time of the path and omitting the number of parking spots along the path leads to a minor decrease in performance (of around 1 to 8). 

Our architecture includes a novel pathing module, which cannot be removed entirely since subsequent modules require a per-action representation. However, we can examine the impact of excluding specific components within this module, namely the action target encoder and the route aggregation module. Our ablation reveals a slight reduction in performance (around 4 to 11 fewer violations fined per day) when leaving out either of these components.

Furthermore, we investigate the impact of removing information that enables the network to differentiate between individual parking spots and learn spot-specific representations. This results in a slight decrease in the average number of parking violations fined (of around 4 to 8).

In summary, our ablation study highlights the importance of the future positioning module in our architecture. Its removal significantly reduces the agent's ability to fine parking violations effectively. Additionally, while we cannot remove the pathing module in its entirety, its individual components, the action target encoder, and the route aggregation module contribute to the performance, although their exclusion has a comparatively smaller impact. Lastly, maintaining the ability to differentiate between individual parking spots and learn spot-specific representation has a minor performance benefit.

\section{Conclusion}
\label{sec:conclusion}

In this paper, we presented SATOP, a novel \ac{RL}-based approach for \ac{TOP} that incorporates various spatial relationships between parking spots, actions, and the officer's location. Our novel architecture consists of a pathing module to learn better representations of actions by including parking spots along each action's path, as well as a future positioning module that allows the agent to assess its positioning after executing an action by learning inter-action correlations. We evaluated our approach and several state-of-the-art baselines using a simulation environment replaying real-world parking data from Melbourne. Our approach consistently outperformed all competitors regarding fined violations, demonstrating its effectiveness in addressing the challenges of \ac{TOP} in realistic scenarios.

Our approach opens up several future research directions.
First, we plan to extend our spatial-aware architecture to the multi-agent version of \ac{TOP}, where the coordination between multiple officers poses a significant challenge.
Additionally, we want to apply and extend our architecture to tackle various other spatial tasks beyond \ac{TOP}. One such example is the \ac{VRP}, which shares some similarities with \ac{TOP} but presents its own set of challenges. 
Finally, we aim to explore scenarios that include dynamically changing routes and travel times due to traffic or other environmental factors.

\bibliographystyle{siam}
\bibliography{sdm}

\end{document}


\newcommand\relatedversion{}

\title{\Large Appendix}

\date{}

\maketitle


\fancyfoot[R]{\scriptsize{Copyright \textcopyright\ 20XX by SIAM\\
Unauthorized reproduction of this article is prohibited}}






\section{Parameters of Environment, Training, and our Architecture}

\begin{table}[h]
    \centering
    \begin{tabular}{|c|c|}
    \hline
         Parameter & Value  \\ \hline

        Officer Speed & 5km/h \\
        Data Year & 2019 \\
        Working Hour Start & 7 \\
        Working Hour End & 19 \\

        \hline

        $\gamma$ & 0.999 \\
        $\epsilon$ min & 0.01 \\
        $\epsilon$-decay & exp \\
        Steps til $\epsilon$-decay start & 10000 \\
        Steps until min $\epsilon$ & 5000000 \\
        Optimizer & RMSProp \\
        Learning Rate & 0.0001 \\
        Alpha & 0.99 \\
        Batch Size & 256 \\
        Number of Parallel Envs & 8 \\
        Replay Buffer Size & 100000 \\
        Reward Transformation & None \\
        Number of Episodes & 40 (=8000000 total env steps)\\
        Env Steps per Episode & 200000 \\
        Start Learning & 10000 \\
        Train Every Env Steps & 32 \\
        Target Update Frequency & 3125 gradient steps \\

        \hline
        Parking Spot Encoder (MLP) & 256 HDIM, 4 Layers, ELU (no act after last), LN  \\
        Parking Spot Rep Dim $d_h$ & 256 \\
        Parking Spot ID Linear Emb Dim ($d_{le}$) & 64 \\
        Action Representation Net ($MLP_{ah}$) & 1024 HDIM, 4 Layers, ELU, LN, 256 Out Dim  \\
        Action Target Hidden Dim ($d_{at}$) & 256 \\
        Future Pos Hidden Dim $d_{ah}$ &  256\\
        Complex Edge Info Net ($MLP_{\delta}$) & 256 HDIM, 2 Layers, Tanh \\
        Action Target Encoder Activation ($\sigma$) & None \\
        Q-Net ($MLP_Q$) & 256 HDIM, 4 Layers, ELU (no act after last), LN  \\
        Norm Complex Edge Info & False,True \\
        Future Pos Activation ($\sigma$) & ELU \\
        Number of Future Pos Layers ($N$) & 2 \\
        Distance Norm Factor & 1/3000.0 \\
        Duration Norm Factor & 1/3000.0 \\
        Route Norm Factor ($\phi_a$) & 1/3000.0 \\
        Learnable Parameter Route ($\theta$) & True \\
        Norm Sim Matrix & False \\
        \hline
    \end{tabular}
    \caption{Parameters of Environment, Training, and our Architecture}
    \label{tab:hyperparameters}
\end{table}

In this section we list the hyperparameters of our approach.

\section{Input Features}
\begin{table}[h]
    \centering
    \begin{tabular}{|c|c|c|}
         \hline
         Feature & Details & Encoding  \\
         \hline
         Current Status & Free, Occupied, In Violation, Fined & One Hot \\
         \hline
         Optimistic In Violation & - & Boolean \\
         \hline
         Time of Day & - & 0 to 1 \\
         \hline
         Walking Time of Officer & - & Normalized \\
         \hline
         Arrival Time of Officer & - & Normalized \\
         \hline
         Distance To Spot & - & Normalized \\
         \hline
         Occupy/Violation Duration & - & -1 to 2\\
         \hline
         X and Y coordinates & - & Normalized \\
         \hline
    \end{tabular}
    \caption{Parking Spot Features}
    \label{tab:features}
\end{table}
We briefly give a tabular description of the input features.

%% file: acronyms.tex
\DeclareAcronym{RL}{
    short = RL,
    long  = reinforcement learning
}

\DeclareAcronym{SRC}{
    short = SRC,
    long  = stochastic resource collection
}

\DeclareAcronym{TOP}{
    short = TOP,
    long  = traveling officer problem
}

\DeclareAcronym{MTOP}{
    short = MTOP,
    long  = multiple traveling officer problem
}

\DeclareAcronym{TSP}{
    short = TSP,
    long  = traveling salesman problem
}

\DeclareAcronym{VRP}{
    short = VRP,
    long  = vehicle routing problem
}

\DeclareAcronym{TDP}{
    short = TDP,
    long  = taxi dispatch problem
}

\DeclareAcronym{SMDP}{
    short = SMDP,
    long  = Semi Markov Decision Process
}
\DeclareAcronym{MDP}{
    short = MDP,
    long  = Markov Decision Process
}

%% file: custom_commands.tex
\DeclareMathOperator*{\argmax}{argmax}

\newcommand{\graph}{G}
\newcommand{\nodeset}{V}
\newcommand{\edgeset}{E}
\newcommand{\resourceset}{P}
\newcommand{\edge}{e}

\newcommand{\resource}{p}
\newcommand{\travelcosts}{C}
\newcommand{\resourcesonedgeset}{PE}
\newcommand{\resourcesonrouteset}{PR}

\newcommand{\state}{s}
\newcommand{\stateset}{S}
\newcommand{\action}{a}
\newcommand{\actionset}{A}

\newcommand{\resourcerep}{h}